\documentclass{vgtc}     

\ifpdf
  \pdfoutput=1\relax                   
  \usepackage{graphicx, cuted, subfig}                
  \DeclareGraphicsExtensions{.pdf,.png,.jpg,.jpeg} 
\else
  \usepackage{graphicx}                
  \DeclareGraphicsExtensions{.eps}     
\fi%

\graphicspath{{figures/}{pictures/}{images/}{./}} 

\usepackage{microtype}                 
\PassOptionsToPackage{warn}{textcomp}  
\usepackage{textcomp}                  
\usepackage{mathptmx}                  
\usepackage{lipsum}
\usepackage{times}                     
\usepackage{amsmath}
\usepackage{cite}                      
\usepackage{graphicx}
\usepackage{tabu}                      
\usepackage{booktabs}                  

\onlineid{0}

\vgtccategory{Research}

\title{Visual Probing and Correction of Object Recognition Models with Interactive user feedback}

\author{Viny Saajan Victor\\ %
        \scriptsize Technical University Kaiserslautern %
\and Pramod Vadiraja\\ %
     \scriptsize Technical University Kaiserslautern %
\and Jan-Tobias Sohns\\ %
     \parbox{1.4in}{\scriptsize \centering Technical University Kaiserslautern}
\and Heike Leitte\\ %
     \parbox{1.4in}{\scriptsize \centering Technical University Kaiserslautern}}

\abstract{With the advent of state-of-the-art machine learning and deep learning technologies, several industries are moving towards the field. Applications of such technologies are highly diverse ranging from natural language processing to computer vision. Object recognition is one such area in the computer vision domain. Although proven to perform with high accuracy, there are still areas where such models can be improved. This is in-fact highly important in real-world use cases like autonomous driving or cancer detection, that are highly sensitive and expect such technologies to have almost no uncertainties. In this paper, we attempt to visualise the uncertainties in object recognition models and propose a correction process via user feedback. We further demonstrate our approach on the data provided by the VAST 2020 Mini-Challenge 2. 
} 

\keywords{Information Visualisation, Uncertainty Visualization, Human Perception, Cognition}

\begin{document}
\maketitle

\firstsection{Introduction}

Computer Vision technologies has made its way to almost every possible field. Object recognition is one such application where given an image to the model, it can identify multiple classes/objects in the image but also localise each object by predicting the coordinates of the bounding box for each object. Although we can train  the model with a huge data set, being able to generalise the model to cater to real world scenarios is highly challenging. Some of the issues that are faced by state-of-the-art object recognition models are: Real time detection, Handling multiple aspect ratios, Limited data, Skewed data.

As discussed above, it is cumbersome to  create a data set that covers all possible variances that can occur in the data. This in-turn poses a problem in generalising the model to cater to new incoming test points. One way this can be addressed is to leverage expert knowledge to guide the model to handle special cases that may occur in the data.
In this paper, we attempt to address this problem by:
\begin{itemize}
\setlength\itemsep{0em}
    \item Proposing a generalised set of techniques that can be applied to any object recognition model.
    \item Evaluate the proposed approach against the YOLO \cite{redmon2016yolo9000} object recognition model, as part of the VAST 2020 IEEE Mini-Challenge 2.

\end{itemize}
 \section{Proposed Approach}
    \begin{figure}
    \centering
    \includegraphics[scale=.15]{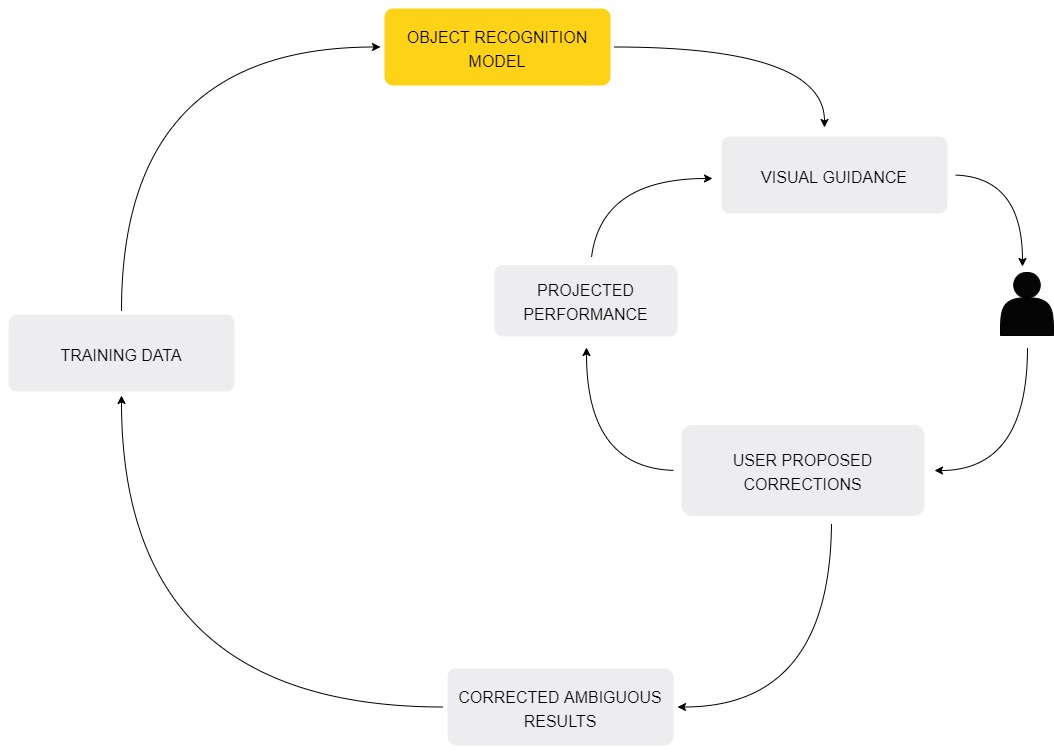}
    \caption{Proposed Approach}
    \label{fig:approach}
\end{figure}
As seen in Figure ~\ref{fig:approach}, we include the expert user in the introspection of the object recognition model. Model's predictions are presented to the user in the form if interactive visualisations . The user then corrects the errors committed by the classifier. These changes in-turn affects the performance of the model. The projected performance of the classifier is made available to the user that guides the user in the correction process. The corrected data is then passed back to the model, thus enabling it to be better prepared for such ambiguities.
We formulate two approaches for associating the model with the user -  Classifier Analysis and Classifier Correction.
\subsubsection{Classifier Analysis}
In this phase, we try to evaluate the performance of classifier by introspecting its predictions. 
We consider the following approaches for analysing the classifier performance:
\begin{itemize}
\setlength\itemsep{0em}
    \item \textbf{Correlating mean and variances of the confidence scores for each object with its average bounding box size}. While it is expected that the objects with larger bounding sizes will have higher confidences, we can leverage this analysis to identify and isolate outlier points and conclude how or why such cases exist by analysing the corresponding images. Using our tool, we concluded that objects with larger bounding boxes have higher confidence scores in comparison with the objects with smaller bounding boxes. This reflects one of the drawback of the YOLO v2 model 
  
    \item \textbf{Correlating the degree of clutteredness of objects in the image with the average confidence score for that image}. We interpret this metric as the average density of objects in the image, and is computed as given in Eqn 1.

\begin{equation}
\resizebox{0.44\textwidth}{!}{$\rho = \frac{\textrm{\# objects predicted}}{(\max({xcoords}) -
     \min({xcoords}) \times (\max(ycoords) - \min({ycoords}))}$}
\end{equation}

    where, $\rho$ is the average density of objects in the a given image, $xcoords$ represents the x-coordinates of the bounding boxes of objects and $ycoords$ represents the y-coordinates of the bounding boxes of objects in that image. On leveraging the analysis made available by our tool, we concluded that the given model performs poorly for the images with highly cluttered objects. This in turn reflects another disadvantage of the YOLO model due to the fact that each grid can propose only 2 bounding boxes in the model.
    
\end{itemize}
\subsubsection{Classifier Correction}
\begin{figure}[ht]
    \centering
    \includegraphics[scale = .14]{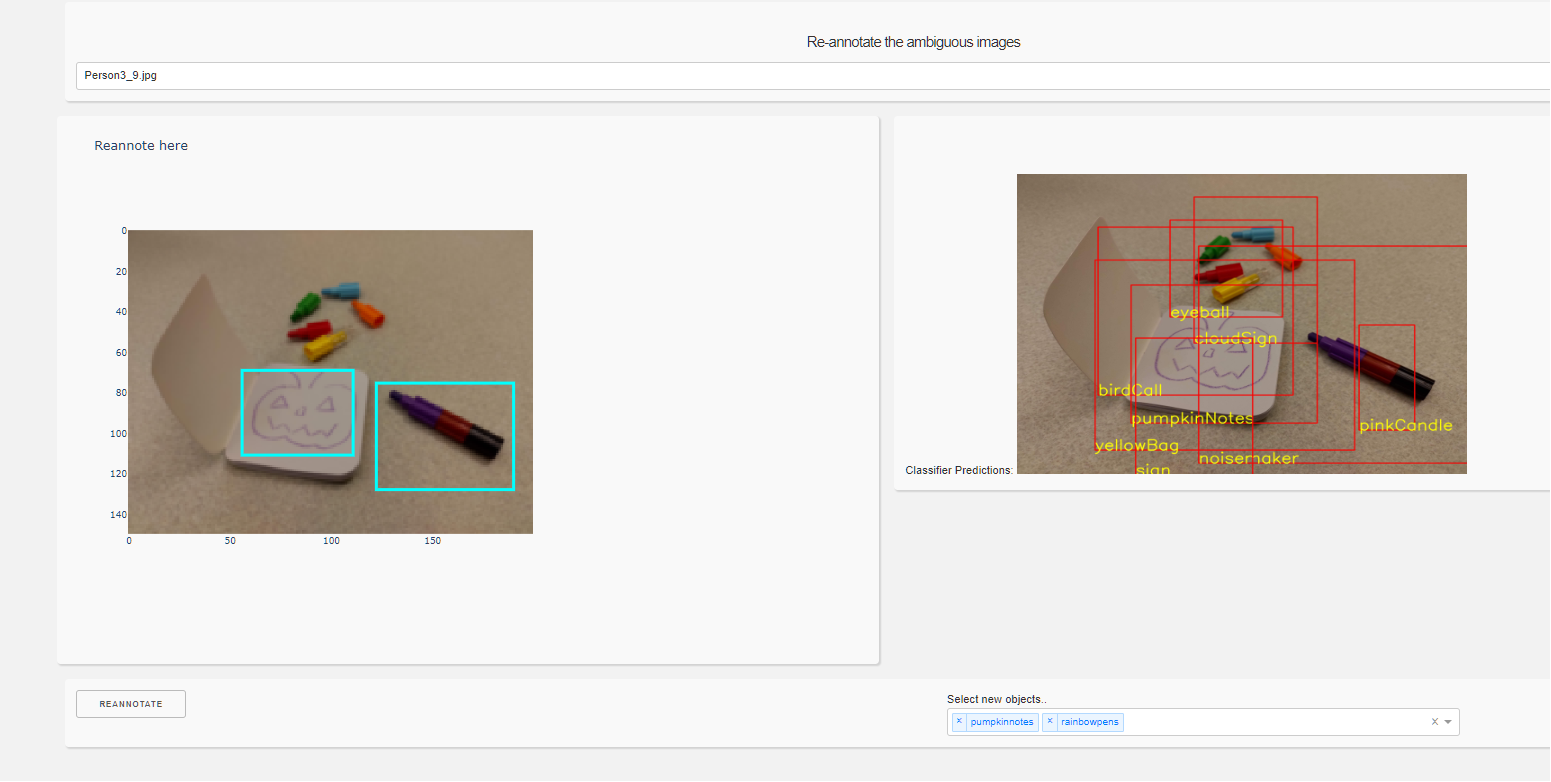}
    \caption{Re-annotating ambiguous images}
    \label{fig:reannotate}
\end{figure}

We consider the following correction mechanisms:
\begin{itemize}
    \item 
    \textbf{Eliminating False Positives through user feedback}:
As the first correction process, we enabled the user inspect the images for which a given object was predicted. This was then assessed by the user to determine its validity and hence treat them as a True or a False Positive. To smoothen the whole process further, we introduced selective attention \cite{selectiveattention} by guiding the user through the process. This was done by showing the user, the objects that are more likely to contain False Positive images using a graph that portrayed proportions of the test images mapped to each object.
\par
Once, the user selects the images to inspect, it is displayed as a grid which in-turn can be selected by the user for elimination. These selected False Positives are treated as flawed results by the classifier, and hence we project how the performance of the classifier improved in terms of the average confidence score for that object. This projection graph is aggregated with its previous changes, and hence gives an overall sense of improvement in the model. This also decreases uncertainty in the model with respect to the object. The entire flow can be seen in Figure \ref{fig:eliminate_fp_graph}. As seen, initially the user chooses a set of images that was mapped to an object. This is followed by successive selections and eliminations of False Positives. During each stage of elimination, a graph records how the performance of the object recognition model with respect to the object is being improved.

\begin{figure}[ht]
    \centering
    \includegraphics[scale = .21]{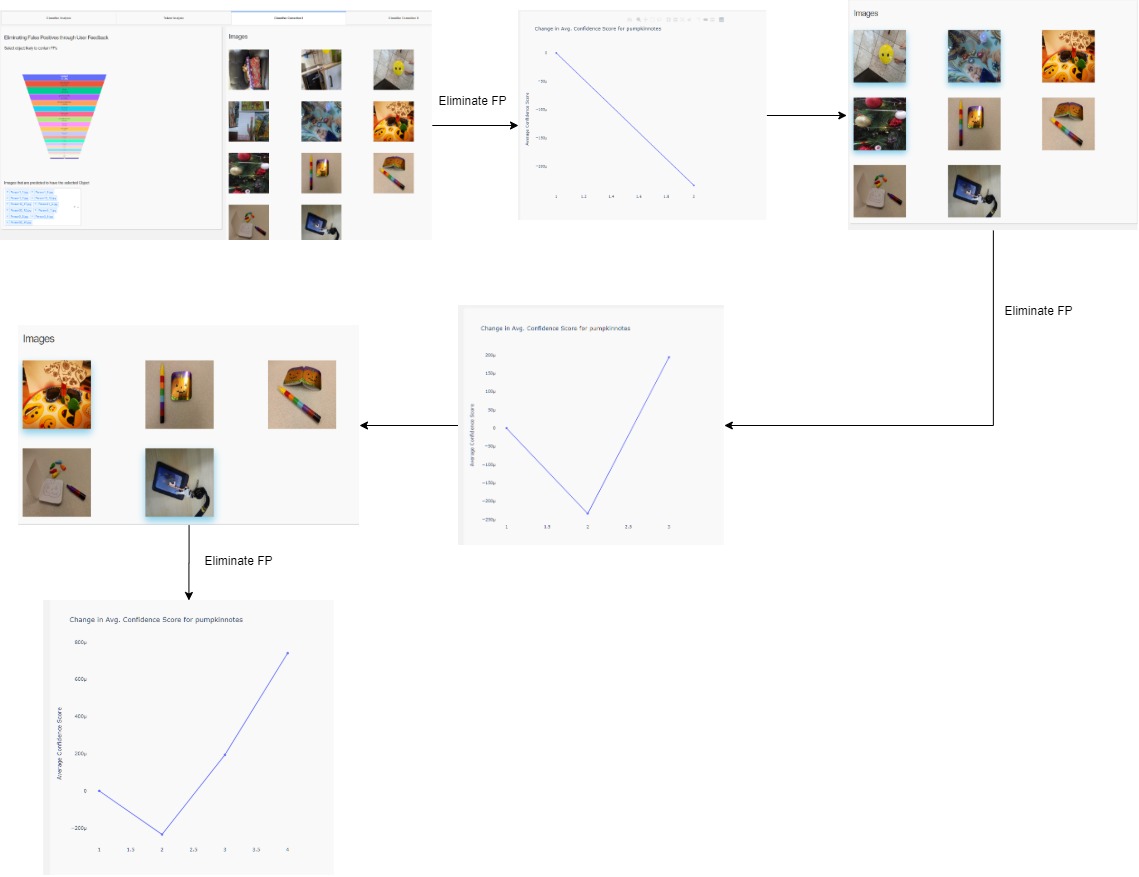}
    \caption{False Positives Elimination Process}
    \label{fig:eliminate_fp_graph}
\end{figure}
\item
\textbf{Re-annotate the ambiguous images}:
For true positives images, there are possibilities that the bounding boxes of the objects were not accurate enough. Hence, we propose an approach as seen in Figure \ref{fig:reannotate} where we allow the user to re-annotate the objects by showing the actual predictions of the bounding boxes to the user. The re-annotated co-ordinates can then be exported by the user that can in-turn be used to update the model to improve its performance.
\item
We present the user,\textbf{ ground-truth vs prediction mapping}(where limited ground-truths were obtained from image captions) and allow the user to re-annotate the ambiguous predictions. The user through inspection can recognise the images with only False Negatives, and with the help of training images, he re annotates the most ambiguous False Negatives with the ground truth.
\end{itemize}

\par
After the above steps of model analysis and correction, the user can move onto the step of finding the totem and the associated group of 8 people.

\section{Totem Analysis}
For this analysis, we relied on the provided captions, prediction labels and text files associated with few persons. The textual content was subjected to the Natural Language Processing pipeline \cite{bird2009natural} that included the steps of tokenization, stop word removal, lemmatization before being considered for further processing. We built a graph structure that connected people based on the common tokens in their captions. These tokens were further filtered out to have only the objects. Hence we had a graph that connected people only if they had common objects.We analyzed this graph by extracting cliques of various sizes.
For the prediction labels, we constructed a similarity matrix that gives the pairwise similarities between the 40 people in terms of the common objects between them. Each person is mapped to a count vector that corresponds to the count of each of the objects. Then we took cosine similarities and colored them based on similarity.
The above two types of analysis along with the user's intervention allowed us to find the resulting totem and the associated group.  

\begin{figure}[ht]
    \centering
    \includegraphics[scale = .20]{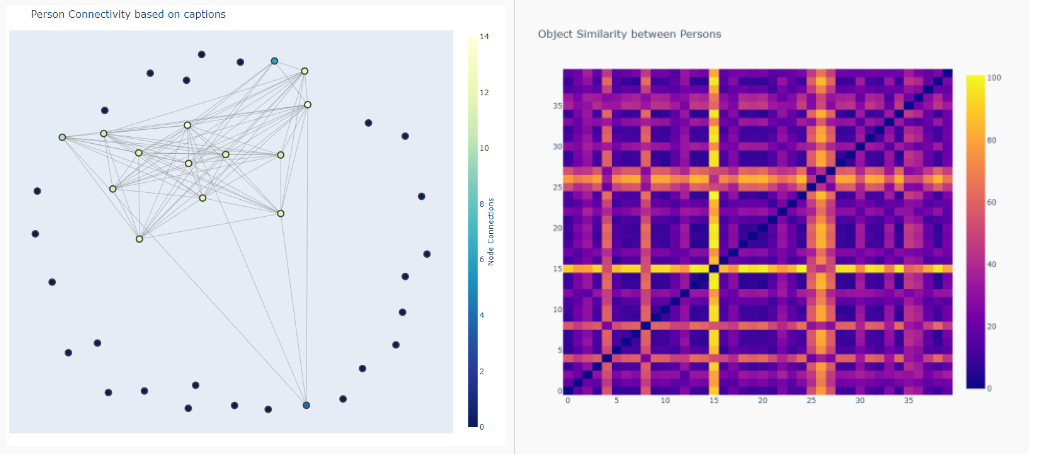}
    \caption{}Network Graph and Similarity Matrix
    \label{fig:net_graph}
\end{figure}

\section{Conclusion}
In this paper, we explored ways of improving the performance of object recognition models. We proposed a set of approaches in the lines of classifier analysis and corrections. We also explored ways of leveraging user perception to humanize the model's perception. These approaches were implemented as part of the VAST 2020 Mini-Challenge2 and respective results were presented.
\bibliographystyle{abbrv-doi}

\bibliography{template}
\end{document}